\documentclass{article}
\usepackage{natbib}
     \usepackage[final]{neurips_2020}

\usepackage[utf8]{inputenc} 
\usepackage[T1]{fontenc}    
\usepackage{hyperref}       
\usepackage{url}            
\usepackage{booktabs}       
\usepackage{amsfonts}       
\usepackage{nicefrac}       
\usepackage{microtype}      
\usepackage{physics}
\usepackage{graphicx}
\UseRawInputEncoding

\title{Deep Spatial Learning with Molecular Vibration}

\author{Ziyang Zhang \\
  National University of Singapore\\
  Singapore 119077 \\
  \texttt{E0546242@u.nus.edu} \\
  \And
  Yingtao Luo \\
  University of Washington\\
  Seattle, WA 98195 \\
  \texttt{yl3851@uw.edu} \\
}

\begin{document}

\maketitle

\begin{abstract}
  Machine learning over-fitting caused by data scarcity greatly limits the application of machine learning for molecules. Due to manufacturing processes difference, big data is not always rendered available through computational chemistry methods for some tasks, causing data scarcity problem for machine learning algorithms. Here we propose to extract the natural features of molecular structures and rationally distort them to augment the data availability. This method allows a machine learning project to leverage the powerful fit of physics-informed augmentation for providing significant boost to predictive accuracy. Successfully verified by the prediction of rejection rate and flux of thin film polyamide nanofiltration membranes, with the relative error dropping from 16.34\% to 6.71\% and the coefficient of determination rising from 0.16 to 0.75, the proposed deep spatial learning with molecular vibration is widely instructive for molecular science. Experimental comparison unequivocally demonstrates its superiority over common learning algorithms.
\end{abstract}

\section{Introduction}

Machine learning powered by big data is an efficient solution to learn data representations of high dimensions and huge complexity. \citet{lake2015human} \citet{butler2018machine} The effectiveness of deep learning depends on the quality and quantity of data. Some systems have existing databases available, such as the Cambridge Crystallographic Data Center for crystallographic graphic classification. \citet{wang2018synchrotron} For other systems lacking the database availability, big data is rendered available through computational chemistry methods like Molecular Dynamics. \citet{ye2019neural} However, an extensive universal database is not readily available with other materials that involve consecutive experiments with too many parameters to calculate on computational software, such as the polyamide thin film composite nanofiltration membranes. While direct calculation is about to consume unnecessarily huge computing resources, our motivation here is to apply few-shot learning methods based on data augmentation to make deep learning effective with limited data sources. 

In thermodynamics, molecular vibration is a natural property to describe the motion of a multi-atom molecule. In 3D space, the exact position of an atom is described by Schrodinger wave equation, which enriches the feature variations for a more accurate modelling. The mechanism of proposed method is similar to the rational distortions used in face perception, such as shifting, scaling, and rotation. These operations can create available data without harming the recognizable features to improve model accuracy, as the rotated or translated human face will still be recognized as the same person. \citet{young2013configurational} The only but non-trivial difference here is that vibration is in line with physical law, which is necessary for interpretable machine learning. Studies on topics such as proton conductivity, \citet{liu2018statistical} quantum force field, \citet{nielsen2018deoxyfluorination} and drug discovery \citet{ekins2019exploiting} extensively use representations such as Simplified molecular input line entry specification (SMILES), graphs, electronic density and etc., but rarely consider to deploy spatial representation. SMILES only has 1D linear information, discarding the spatial information; Chemical compound formula does not present information like bond length and bond angle; Electronic density is an option with detailed monomer information, but not as intuitive as geometry. Spatial representation, on the other hand, does not discard any information. Hereby, it puts forward an interest whether representation learning based on spatial distribution can also be effective for molecule science. We practiced spatial representation and molecular vibration to tackle the deep learning of polyamide nanofiltration membranes along with the data scarcity problem. In seeking a result efficient workflow, we performed comparative experiments on applicable data set size and machine learning algorithm. Experimental results have unequivocally shown its advantage in boosting accuracy. 

\begin{figure}
  \centering
  \includegraphics[width=14cm, height=9.5cm]{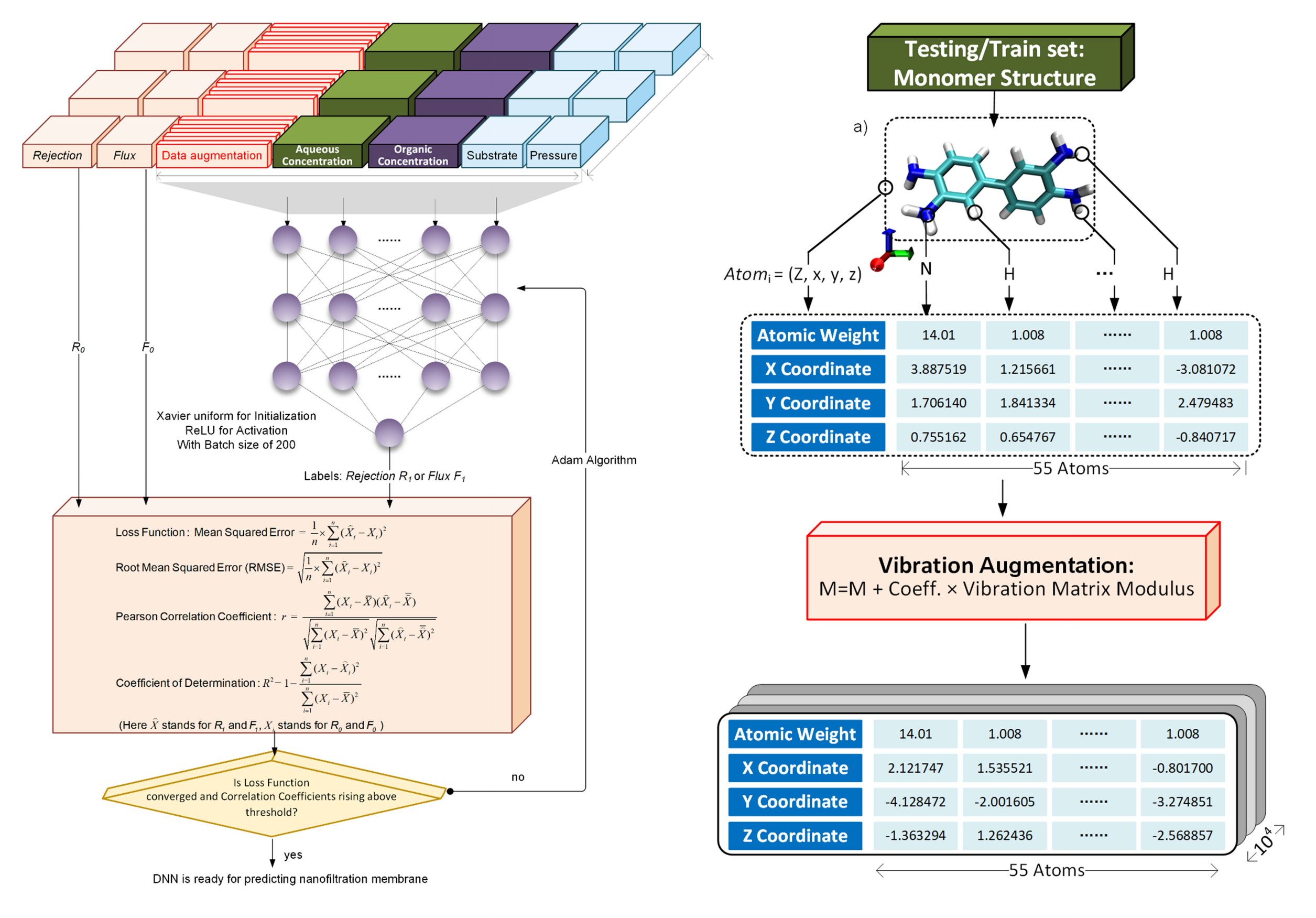}
  \caption{Diagram of deep spatial learning with vibration augmentation. Supporting data augmentations of Molecular Vibration for Cartesian coordinates are exemplified on 3,3'-Diaminobenzidine. We use augmentation to create datasets of sizes from $10^2$ to up to $10^4$. As neural network converged, (Root) Mean Squared Error (MSE/RMSE), Pearson Correlation Coefficient (PCC), Mean Relative Error (MRE), Coefficient of Determination (R2) are used as metrics to measure its performance.}
\end{figure}

\section{Experimental}
In literature, polyamide TFCMs were prepared by interfacial polymerization under varied conditions \citet{guo2020high} \citet{xiao2019amphibian} \citet{sheng2020electro}. Among them, four variables determining the nanofiltration performance were identified in this work, I: substrate membranes; II: monomer structures; III: monomer concentrations; IV operation pressure. For data collection, nanofiltration membranes are indexed in Web of Science in line with the following criteria: only one monomer (acyl chloride/amine monomers) was dissolved in the aqueous phase and the organic phase (hexane); supporting membranes were limited to polysulfide, polyethersulfone and polyacrylonitrile. Salt rejection of TFCMs is represented by NaSO4 rejection. The monomer molecules are optimized at B3LYP-D3/6-311+G(d,p) levels using the Gaussian16 program \citet{frisch2016gaussian}. The interfacial polymerization is introduced to 444 descriptor variables, including 2 concentrations of organic phase and aqueous phase, 1 pressure during a nanofiltration examination, 1 substrate's type classification and 440 structural parameters containing 55 atoms of 2 monomers in atomic weight and 3D coordinates (Fig 1a). Here, the 3D coordinates are varied along with molecular vibration.

The raw database (100) collected from previous literature was divided into a training dataset (70/100) and a test dataset (30/100). Deep learning can automatically update its weight by decreasing the loss function (MSE) until it can give prediction values (outputs, i.e., salt rejection and flux) close enough to the real values (labels). Four metrices are used for evaluation: Pearson correlation coefficient (PCC), Mean Relative Error (MRE, \%), Root Mean Squared Error (RMSE) and Coefficient of Determination $R^2$. The training and test datasets are defined for seperate use, while the test data was not fed to update DNN's weights and bias. Hence, the validity of our method can be evaluated by the results (MSE, RMSE, MRE, PCC, R2) of the test dataset. As mentioned above, data preparation considers 440 spatial distribution along with atomic number. For small molecules of less than 55 atoms, zeros are padded to their ends. As our work focuses on the predictive improvement of this physics-informed vibration, as shown in Fig 1, we choose fully connected layers as the sole module and leave topics about how to employ this technique in different neural architectures with other representations to future studies. The number of neurons of each layer in our model was set to 444-100-20-1. Despite the simplicity of baseline neural network, we show in the next section that it outperforms popular machine learning algorithms that are commonly practiced by recent studies, such as random forests.

According to quantum chemistry, vibration matrix is randomly selected after structure optimization calculation, and the maximum vibration amplitude of the vibration is calculated by the force constant under a certain environmental energy, obeying the normal distribution centred on the original position. Further details and implementation codes regarding the vibration technique is discussed in \textit{https://github.com/yingtaoluo/Nanofiltration-Membrane-Deep-Learning}.

\begin{figure}
  \centering
  \includegraphics[width=14cm, height=5.5cm]{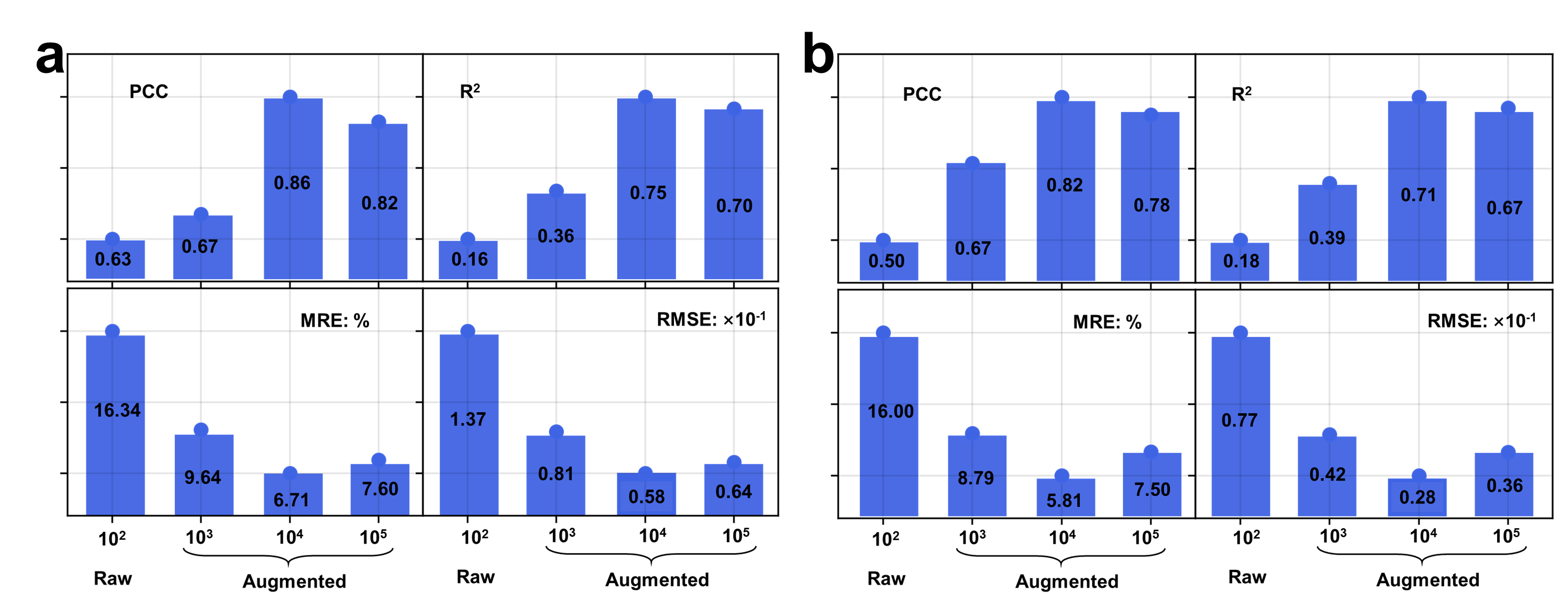}
  \caption{Deep learning's performance on test set with different augmentations in spatial representation. Test [PCC, R2, MRE and RMSE] of [raw $10^2$ dataset, augmented $10^3$, $10^4$ and $10^5$] dataset on a) rejection and b) flux with the 1st notion. This experiment shows that augmentation boosts all indicators by a large margin. Please note: for comparison, all hyper-parameters were set to the same to ensure fairness at training.}
\end{figure}

\section{Results}
In Fig 2, we show the positive impact of our molecular augmentation on deep spatial representation learning by comparing the error (RMSE and MRE) and the correlation degree (PCC and R2) on test sets between raw and augmented data of different sizes. Both raw and augmented dataset are evaluated with the same hyper-parameters during the training process to make a fair comparison, excluding the effect of training strategies. Both errors and correlations on test dataset have been improved as a result of molecular augmentation in the deep spatial representation. The best performance comes with the augmented $10^4$-sized dataset, with the PCC rises from 0.63/0.5 to 0.86/0.82, the R2 rises from 0.16/0.18 to 0.75/0.71, the MRE drops from 16.34\%/16\% to 6.71\%/5.81\%, and the RMSE drops from 0.137/0.077 to 0.058/0.028. PCC and R2 above 0.7 suggest strong correlation between predictions and labels, and MRE below 10\% means a very accurate prediction to guide fabrication.

From the experiments, we find that vibration method is always helpful, although augmentation $10^5$ generates too much data that training becomes tricky in a fixed hyper-parameters scheme. This boost is in line with statistics, as vibration smooths the solution space and fills in the blank to alleviate deep learning models' burden in giving a good prediction where training data is not covered. Both raw and augmented loss on training data can converge to a minimum, but further generalization to uncovered areas requires an exploration of the solution space. This vibration gives such an exploration that is in line with the chemical rule in determining molecule positions. The size $10^4$ is the best trade-off between accuracy and computational complexity of data generation and training. Compared to actual collection of new experiemental data that may cost months, a 100x augmentation costs less than a day on a normal CPU even for the most complicated molecule known in polyamide membranes.

Fig 3 shows that deep learning with augmented dataset has an advantage over conventional machine learning models. Models like Supporting Vector Regression (SVR) and Gradient Boosting Regression (GBR) tend to be absolute failures as their PCCs are zeros and $R^2$s are negative, which suggests that some conventional algorithms are brittle in a limited data scheme. Random Forest (RF), as one of the most effective machine learning algorithms, can beat DNN with non-augmented $10^2$ raw data. However, the proposed approach that incorporates the deep spatial learning with molecular augmentation leverages the post-augmentation improvement to the maximum and finally wins over RF. All these show that even in the most extreme condition where very little raw data can be collected, the proposed approach can still give us valid and accurate predictions to be used for material fabrication. The very accurate predictions provided by deep spartial representation learning can help filter the membranes that are prepared to be fabricated and avoid the trial-and-error procedure, making material on-demand design much more efficient. It is for the first time in the membrane community to report deep learning instructive with limited data, and experiments on more materials can further verified the generalized effectiveness of the proposed method.

\begin{figure}
  \centering
  \includegraphics[width=14cm, height=5.5cm]{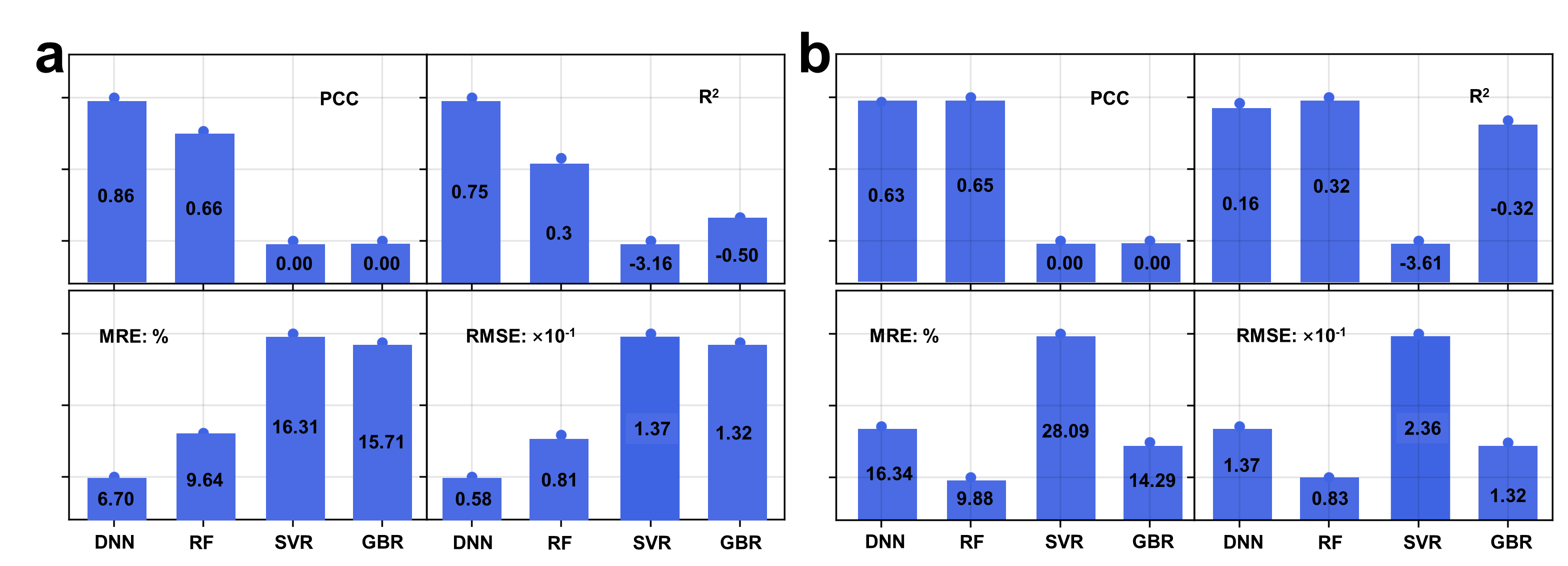}
  \caption{Comparison of different machine learning models on test set. Comparison of different machine learning models on test set. Test [PCC, MRE, RMSE and R2] of [Deep Neural Network (DNN), Random Forest (RF), Supporting Vector Regression (SVR) and Gradient Boosting Regression (GBR)] on rejection prediction in spatial representation with a) augmented and b) raw dataset. This experiment shows that RF has the best results with raw dataset and DNN performs the best with augmented dataset. The overall best result comes from deep learning with molecular augmentation.}
\end{figure}

\section{Conclusion}
We proposed the idea of using deep spatial representation learning and molecular vibration to model relation between laboratory preparing conditions and performances of polyamide nanofiltration membranes. This method shows an approach using an end-to-end manner to exploit the physical information contained in dataset. Molecular vibration as a physically interpretable method provides a significant performance boost to deep learning models, and we demonstrate empirically that this augmentation can empower deep learning to be robust. We prove strong correlation and low error between prediction values of the neural network and real values of the data label, demonstrating the generalization ability of the our model under a limited data scheme. Without expensive computation, the proposed method can open the door to the marriage of machine learning and many molecular materials, to address data scarcity for machine learning applications. Future works upon this technique, such as proposing a simple algorithmic approximation of vibration, or utilizing this technique and many other physics-informed augmentation in other schemes can further push forward this research. 

\small
\bibliography{ref}
\end{document}